\documentclass[sigconf]{acmart}
\AtBeginDocument{%
  }

\setcopyright{acmlicensed}
\copyrightyear{2026}
\acmYear{2026}
\acmDOI{XXXXXXX.XXXXXXX}
\acmConference[SIGIR '26]{Proceedings of the 49th International ACM SIGIR Conference on Research and Development in Information Retrieval}{July 20--24, 2026}{Melbourne, Australia}
\acmISBN{978-1-4503-XXXX-X/2018/06}

\usepackage{multirow}
\usepackage{pifont}
\usepackage{adjustbox}
\usepackage{array}

\begin{document}

\title{VQPP: Video Query Performance Prediction Benchmark}

\author{Adrian Catalin Lutu}
\affiliation{%
  \institution{University of Bucharest / Bitdefender}
  \city{Bucharest}
  \country{Romania}
}
\email{alutu@bitdefender.com}

\author{Eduard Poesina}
\affiliation{%
  \institution{University of Bucharest}
  \city{Bucharest}
  \country{Romania}
  }
\email{eduardgabriel.poe@gmail.com}

\author{Radu Tudor Ionescu}
\authornote{Corresponding author.}
\affiliation{%
  \institution{University of Bucharest}
  \city{Bucharest}
  \country{Romania}
}
\email{raducu.ionescu@gmail.com}

\renewcommand{\shortauthors}{Lutu et al.}

\begin{abstract}
Query performance prediction (QPP) is an important and actively studied information retrieval task, having various applications, such as query reformulation, query expansion, and retrieval system selection, among many others. The task has been primarily studied in the context of text and image retrieval, whereas QPP for content-based video retrieval (CBVR) remains largely underexplored. To this end, we propose the first benchmark for video query performance prediction (VQPP), comprising two text-to-video retrieval datasets and two CBVR systems, respectively. 
VQPP contains a total of 56K text queries and 51K videos, and comes with official training, validation and test splits, fostering direct comparisons and reproducible results. We explore multiple pre-retrieval and post-retrieval performance predictors, creating a representative benchmark for future exploration of QPP in the video domain. Our results show that pre-retrieval predictors obtain competitive performance, enabling applications before performing the retrieval step. We also demonstrate the applicability of VQPP by employing the best performing pre-retrieval predictor as reward model for training a large language model (LLM) on the query reformulation task via direct preference optimization (DPO). We release our benchmark and code at \url{https://github.com/AdrianLutu/VQPP}.
\end{abstract}


\begin{teaserfigure}
\centering
  \includegraphics[width=0.8\textwidth]{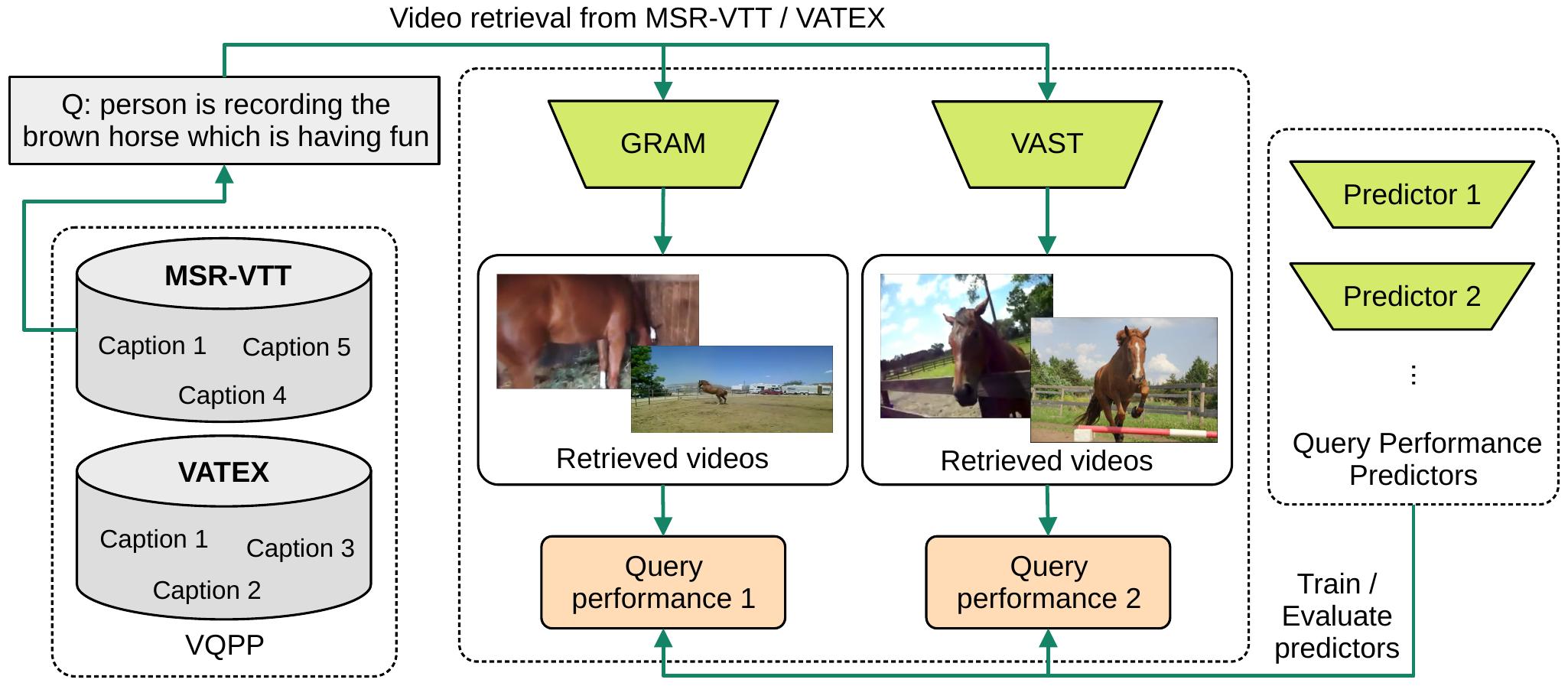}
  \vspace{-0.3cm}
  \caption{The Video Query Performance Prediction (VQPP) benchmark comprises two content-based video retrieval datasets: MSR-VTT \cite{xu2016msrvtt} and VATEX \cite{Wang_2019_ICCV}. Queries (captions) from both datasets are given as input to two video retrieval models: GRAM \cite{cicchetti2025gramian} and VAST \cite{Chen-NEURIPS-2023}. Retrieval performance is measured for each query given to each model. Multiple query performance predictors are trained and evaluated on our official splits. Best viewed in color.}
  \vspace{0.1cm}
  \Description{Enjoying the baseball game from the third-base
  seats. Ichiro Suzuki preparing to bat.}
  \label{fig:teaser}
\end{teaserfigure}

\received{20 February 2007}
\received[revised]{12 March 2009}
\received[accepted]{5 June 2009}

\maketitle

\section{Introduction}

One of the challenging problems in information retrieval is query performance prediction (QPP), the task of predicting the retrieval performance of a system for a given query, without ground-truth relevance judgments \cite{CronenTownsend-SIGIR-2002}. Solving this task paves the pathway towards a wide range of applications, including automatic query reformulation \cite{Poesina-CVPR-2025}, automatic query expansion \cite{Rudinac-IJMIR-2012}, retrieval system selection \cite{Poesina-CVPR-2025}, adaptive filtering, etc. Due to its importance, QPP, also referred to as query difficulty estimation, has been actively studied in information retrieval \cite{Arabzadeh-CIKM-2021,Arabzadeh-ECIR-2021,Vinay-SIGIR-2006,Datta-SIGIR-2022,Datta-WSDM-2022,Datta-TIS-2022,Faggioli-IRJ-2022,Jafarzadeh-IPM-2022,Sarwar-ADCS-2021,Meng-ECIR-2025}. However, the mainstream focus of QPP research has been dedicated to text retrieval \cite{CronenTownsend-SIGIR-2002,He-SPIRE-2004,Mothe-SIGIR-2005,Hauff-CIKM-2008,Hauff-ECIR-2009,Shtok-SIGIR-2010,Cummins-SIGIR-2011,Kurland-CIKM-2012a,Kurland-CIKM-2012b,Shtok-TIS-2012,Cummins-TIS-2014,Katz-SIGIR-2014,Raiber-SIGIR-2014,Tao-CIKM-2014,Shtok-TIS-2016,Roitman-SIGIR-2017a,Roitman-SIGIR-2017b,Chifu-SIGIR-2018,Mizzaro-SIGIR-2018,Roitman-SIGIR-2018,Zamani-SIGIR-2018,Roy-IPM-2019,Zendel-SIGIR-2019,Arabzadeh-IPM-2020,Dejean-SAC-2020}. Recently, the interest towards QPP in content-based image retrieval has been growing \cite{Xing-ECIR-2010,Nie-TIS-2012,Tian-TMM-2012,Jia-ICME-2014,Jia-SP-2015,Tian-TMM-2015,Pedronette-SIBGRAPI-2015,Sun-TIP-2018,Li-NC-2012,Valem-ICMR-2021}, with a number of studies, such as iQPP \cite{Poesina-SIGIR-2023} and PQPP \cite{Poesina-CVPR-2025}, trying to stimulate research in this area by providing organized evaluation benchmarks. Yet, QPP for content-based video retrieval (CBVR) remains largely unexplored. To the best of our knowledge, there are only two studies in this direction \cite{Kofler-TMM-2014,Rudinac-IJMIR-2012}, both being conducted more than 10 years ago. Only one of these works actually addressed QPP \cite{Rudinac-IJMIR-2012}, the other being focused on a less challenging task, that of detecting failing queries \cite{Kofler-TMM-2014}. 

While Meng et al.~\cite{Meng-ECIR-2025} recognize the importance of studying video QPP, there is no prior work that ever attempted to develop a QPP benchmark for CBVR. To bridge this gap, we propose the first benchmark for \textbf{v}ideo \textbf{q}uery \textbf{p}erformance \textbf{p}rediction (VQPP), comprising 56K text queries and 51K videos gathered from two CBVR datasets, namely MSR-VTT \cite{xu2016msrvtt} and VATEX \cite{Wang_2019_ICCV}. As illustrated in Figure \ref{fig:teaser}, ground-truth query performance is assessed with respect to the videos retrieved by two state-of-the-art systems, namely GRAM \cite{cicchetti2025gramian} and VAST \cite{Chen-NEURIPS-2023}. Therefore, our benchmark enables the exploration of QPP across distinct query and video collections, as well as distinct retrieval systems. More precisely, VQPP comprises four evaluation scenarios (two datasets $\times$ two retrieval systems) that can expose the generalization capacity of query performance predictors. Aside from establishing evaluation scenarios, we evaluate multiple pre-retrieval and post-retrieval predictors, ranging from simple linguistic features to deep vision-language models, thus obtaining a comprehensive benchmark. Interestingly, we find that deep pre-retrieval predictors obtain competitive performance against post-retrieval predictors. In fact, the fine-tuned BERT \cite{Devlin-NAACL-2019} pre-retrieval predictor yields top scores across all evaluation scenarios and correlation measures. 

Going beyond the new benchmark, we showcase an application of the fine-tuned BERT predictor in query reformulation. More precisely, we use the fine-tuned BERT predictor as reward model in a preference learning setup where we attempt to fine-tune the Phi-4-mini-instruct language model \cite{Abouelenin-Arxiv-2025} to reformulate queries. We employ Direct Preference Optimization (DPO) \cite{Rafailov-NeurIPS-2023} over pairs of reformulated queries, in which the winning and losing reformulations are established via the fine-tuned BERT predictor. Our results show that using the Phi-4-mini-instruct model to reformulate queries improves retrieval performance.

In summary, our contribution is threefold:
\begin{itemize}
    \item We develop the first benchmark for query performance prediction in content-based video retrieval.
    \item We present extensive experiments with various pre-retrieval and post-retrieval predictors over the four evaluation scenarios included in our benchmark.
    \item We harness the best predictor as a reward model to train a large language model for query reformulation.
\end{itemize}

\section{Related Work}


\noindent\textbf{Query performance predictors.}
The motivation behind early QPP research was to develop systems capable of assisting users to improve retrieval results for a given query \cite{Losee-IPM-1991}. As the field matured, large-scale evaluations \cite{voorhees-ARXIV-2003,clarke-TREC-2009} uncovered substantial variability in system effectiveness across queries. This led to establishing QPP as a research direction in its own right. Over time, a categorization of QPP methods has emerged, dividing predictors into pre-retrieval and post-retrieval. Pre-retrieval predictors operate before any search is performed, relying on features extracted from the query and the document collection: query length, inverse document frequency \cite{He-SPIRE-2004}, linguistic properties of the query \cite{Mothe-SIGIR-2005}, coherence based methods \cite{he-ecir-2008}, and similarity/variability evidence \cite{Zhao-ECIR-2008}. In contrast, post-retrieval predictors operate after one or more retrieval runs have been executed, analyzing the ranked result list, the associated retrieval scores, and the distribution of documents in the output, through measures such as query clarity score \cite{CronenTownsend-SIGIR-2002}, query feedback \cite{Zhou-SIGIR-2007}, query-drift estimation \cite{Shtok-TIS-2012}, clustering tendency, document perturbation and query perturbation \cite{Vinay-SIGIR-2006}.


With the paradigm shift towards neural networks in information retrieval, QPP methods have evolved accordingly, moving from hand-crafted statistical signals to learned representations \cite{Arabzadeh-CIKM-2021, Marcus-PVLDB-2019, Zamani-SIGIR-2018}. Seminal studies in this direction are NeuralQPP \cite{Zamani-SIGIR-2018} and BERT-QPP \cite{Arabzadeh-CIKM-2021}.
More recently, large language models (LLMs) have been integrated into QPP pipelines. For instance, QPP-GenRE \cite{meng-tois-2025} is a framework that uses LLMs to generate pseudo-relevance judgments for the retrieved ranked list, then derives performance estimates by computing standard evaluation measures from these labels. 

Beyond text-based ad-hoc retrieval, QPP has attracted increasing attention in image and video retrieval \cite{Xing-ECIR-2010,Nie-TIS-2012,Jia-SP-2015,Sun-TIP-2018,Kofler-TMM-2014,Rudinac-IJMIR-2012}. In text-to-image search, Xing et al.~\cite{Xing-ECIR-2010} addressed query difficulty prediction for image retrieval, learning classification models based on linguistic features of the query such as concreteness, commonness, and ambiguity. 
In content-based image retrieval, where the query itself is an image, Sun et al.~\cite{Sun-TIP-2018} proposed to evaluate retrieval quality from the top-ranked results by generating a pairwise correlation matrix and training a CNN regression model. In content-based video retrieval, Rudinac et al.~\cite{Rudinac-IJMIR-2012} integrated QPP into a data-driven query expansion framework, while Kofler et al.~\cite{Kofler-TMM-2014} only attempted to detect failing queries. To the best of our knowledge, our study is the first to perform an extensive evaluation of predictors in CBVR.

\noindent\textbf{Query performance prediction benchmarks.}
The maturation of QPP as a research area has been closely tied to the availability of benchmarks that enable reproducible comparisons across methods. The TREC Robust Track \cite{voorhees-ARXIV-2003} provided one of the earliest and most widely used datasets. The TREC Web Track \cite{clarke-TREC-2009} extended evaluation to web-scale collections. More recently, the TREC Deep Learning Track \cite{craswell-TREC-2025}, built on MS MARCO \cite{nguyen-ARXIV-2016}, has become the dominant dataset for QPP research, providing large-scale query sets with relevance judgments. Poesina et al.~\cite{Poesina-SIGIR-2023} introduced the first benchmark for image query performance prediction in the query-by-example setting. More recently, QPP concepts have been extended to generative models by Bizzozero et al.~\cite{bizzozzero2023prompt} and Poesina et al. \cite{Poesina-CVPR-2025}. 
Despite the breadth of QPP research across text retrieval, image search, and image generation, query performance prediction for video retrieval remains largely unexplored. Text-to-video retrieval poses unique challenges compared with text and image retrieval, including the temporal dimension of video content, the multimodal nature of video representations, and the increased computational cost of retrieval over large video collections. To the best of our knowledge, no prior work has established a benchmark specifically designed for QPP in the video domain.

\section{VQPP Benchmark}

\subsection{Overview}

Video Query Performance Prediction (VQPP) is a novel benchmark that establishes a standardized evaluation protocol for predicting the effectiveness of content-based retrieval models at the query level. VQPP aggregates performance data across 56K queries and 51K videos, providing a comprehensive benchmark derived from two diverse video datasets, MSR-VTT \cite{xu2016msrvtt} and VATEX \cite{Wang_2019_ICCV}. By using two retrieval systems, namely GRAM \cite{cicchetti2025gramian} and VAST\cite{Chen-NEURIPS-2023}, VQPP enables the rigorous assessment of QPP estimators in four distinct evaluation scenarios (two datasets $\times$ two retrieval systems).

\subsection{Datasets}

The VQPP benchmark is constructed on top of the MSR-VTT \cite{xu2016msrvtt} and VATEX \cite{Wang_2019_ICCV} datasets. These datasets are selected to enforce diversity in both visual information and captioning style.

\noindent\textbf{MSR-VTT.} The MSR-VTT \cite{xu2016msrvtt} dataset contains 10,000 video clips spanning 20 broad categories, such as music, sports, and news. It represents an ``in-the-wild'' open-domain retrieval setting characterized by significant variance in visual quality, camera motion, and semantic context. All 10,000 videos are included in VQPP. From the total number of captions (20 per video), we randomly select a subset of 32,732 queries.
 
\noindent\textbf{VATEX.} The VATEX \cite{Wang_2019_ICCV} corpus is characterized by shorter video clips, each of around 10 seconds in length. For this benchmark, we exclusively utilize the English captions, to avoid using translated content. The dataset contains 41,250 videos, all of them being imported into VQPP. Each video comes with 10 captions. From the total number of captions, we select a subset of 18,055 queries.

\noindent\textbf{Overall.}
Our benchmark comprises over 51K videos and 56K queries. The resulting number of queries is about two orders of magnitude higher than what is commonly used in text QPP evaluation \cite{meng2023query,Dejean-SAC-2020}, where researchers typically rely on hundreds of queries.



\subsection{Retrieval Systems}

Query performance is typically estimated with respect to a particular retrieval system. To broaden our evaluation setup, we include two state-of-the-art CBVR systems, namely GRAM \cite{cicchetti2025gramian} and VAST \cite{Chen-NEURIPS-2023}. This prevents the benchmark from being biased toward a single model, ensuring that evaluated QPP methods generalize across distinct retrieval models.

\noindent\textbf{GRAM.} Cicchetti et al.~\cite{cicchetti2025gramian} proposed a video retrieval model called GRAM. They introduced a novel objective function that minimizes the Gramian volume of the parallelotope created by modality vectors. Unlike traditional contrastive approaches that rely solely on pairwise cosine similarity, GRAM enforces a stricter geometric structure in the high-dimensional embedding space, resulting in superior fine-grained alignment between text and visual features.
    
\noindent\textbf{VAST.} Chen et al.~\cite{Chen-NEURIPS-2023} introduced VAST, a foundation model trained to integrate auxiliary modalities alongside visual frames. By integrating multimodal information from audio tracks, video clips, subtitles and caption, VAST becomes robust in scenarios where visual cues are ambiguous, but semantic information is present in the other modalities, e.g.~the audio track.
    
\subsection{Organization}

The VQPP benchmark is designed to decouple the QPP task from the computationally expensive video retrieval process. By providing pre-computed retrieval results and performance scores, we enable the development and evaluation of QPP estimators without having to run the heavy retrieval systems or use raw video data. 
    
To support the development of supervised QPP estimators, we provide dense performance annotations for every query-model pair. Video retrieval was executed on the full set of 56K queries for both systems, generating approximately 112K unique performance data points. Data is stratified by model, dataset and split. Each entry contains the textual query and a vector of performance scores, corresponding to different evaluation measures, namely Reciprocal Rank (RR) and Recall@$K$ ($K \in \{1, 5, 10, 20\}$). This format allows for the immediate training of QPP models. To enable the development of post-retrieval QPP methods, which typically analyze the score distribution or coherence of retrieved videos, we release the top-100 retrieval results for all 112K query-model pairs. This eliminates the need for running video retrieval again, allowing researchers to compute standard predictors or extract coherence features directly from the candidate lists.

We partition the queries in each dataset into distinct train, validation, and test subsets. In Table \ref{tab:splits}, we present the number of videos and the number of queries per split, for each of the two datasets. Most queries are reserved for training, while keeping a sizable amount of queries for testing to enable reliable query performance predictor benchmarking. 

\begin{table}[t]
    \centering
    \caption{Statistics of the VQPP benchmark. For each dataset, we report the total number of source videos and the distribution of queries across our training, validation, and test splits.}
    \label{tab:splits}
    \vspace{-0.25cm}
    \begin{tabular}{l  c  r r r}
        \toprule
        \multirow{2.5}{*}{\textbf{Dataset}} & \multirow{2.5}{*}{\textbf{\# Videos}} & \multicolumn{3}{c}{\textbf{\# Queries per Split}} \\
        \cmidrule(lr){3-5}
         & & \multicolumn{1}{c}{\textbf{Train}} & \multicolumn{1}{c}{\textbf{Val}} & \multicolumn{1}{c}{\textbf{Test}} \\
        \midrule
        MSR-VTT & 10,000 & 24,490 & 879 & 7,374 \\
        VATEX   & 41,250 & 13,872 & 714 & 3,469 \\
        \bottomrule
    \end{tabular}
    \vspace{-0.2cm}
\end{table}

\subsection{Evaluation Measures}

We employ distinct evaluation measures to assess retrieval performance (this is what query performance predictors aim to predict) and QPP performance (this is how the effectiveness of query performance predictors is determined), respectively.

\noindent\textbf{Retrieval performance.}
Both MSR-VTT and VATEX utilize a one-to-one query-video mapping, i.e.~there is only one correct video for each query. To establish the ground-truth difficulty for each query, we utilize two information retrieval metrics suitable for the one-to-one mapping scenario, namely Reciprocal Rank and Recall@10. The \emph{Reciprocal Rank} (RR) is calculated as the inverse of the rank of the correct video ($1/r$, where $r$ is the rank of the correct video). For each query, \emph{Recall@10} is a binary indicator measuring whether the only target video appears within the top-10 retrieved results or not. This reflects the system's ability to place the correct result within a typical user's attention window.

\noindent\textbf{QPP performance.}
Following conventional evaluation protocols in text~\citep{yom2005learning,Zhao-ECIR-2008} and image~\citep{Poesina-SIGIR-2023,Poesina-CVPR-2025} QPP, we evaluate performance predictors in terms of the Pearson $\rho$ and Kendall $\tau$ correlation coefficients between the predicted and the ground-truth performance levels. \emph{Pearson $\rho$} assesses the linear relationship between predicted difficulty and actual performance.  A high Pearson correlation indicates that the predictor accurately tracks the magnitude of the retrieval metric. \emph{Kendall $\tau$} measures the rank correlation between the lists. This is crucial for QPP, as it quantifies the ability to correctly distinguish between hard and easy queries relative to one another, regardless of the absolute scale of the predicted scores. Furthermore, we test the significance of the results with respect to the random chance baseline using Student's t-testing~\citep{Roitman-SIGIR-2018}.

\section{Predictors}

We implement and evaluate a range of query performance predictors, categorized into pre-retrieval and post-retrieval methods, based on whether they require the search results to generate a prediction or not.

\subsection{Pre-Retrieval Predictors}

Pre-retrieval predictors estimate difficulty using only the query itself, making them highly efficient and suitable for many practical applications, as they do not require interrogating the retrieval system with the given query.

\noindent
\textbf{Linguistic baselines.} We use several statistical heuristics to analyze the surface properties of the query text. These include \textit{synset counts} to measure word ambiguity, \textit{word count} for query length, and average counts of specific part-of-speech tags, such as \textit{numerals}, \textit{conjunctions}, and \textit{prepositions}.
    
\noindent 
\textbf{Fine-tuned BERT.} We train a regression model using a BERT-base-cased \cite{Devlin-NAACL-2019} backbone. We attach a regression head to the \texttt{[CLS]} token returned by the last convolutional block. The model takes the textual query as input and is trained to directly predict the expected Reciprocal Rank or Recall@10. This allows the predictor to capture semantic nuances that simple linguistic counts might miss.
    
\noindent 
\textbf{Few-shot Llama-3.1.} We leverage the open-source Llama-3.1-8B \cite{grattafiori2024llama} large language model (LLM) for query performance prediction in a few-shot prompting setup. We create a prompt in which we provide 16 illustrative examples of queries and their corresponding performance scores. The chosen examples are custom to each test query. More precisely, we apply a k-NN model based on pre-trained BERT embeddings \cite{Devlin-NAACL-2019} to select 16 query examples from the training set. The LLM is then asked to predict the difficulty of the test query based on in-context learning.

\begin{table*}[t]
    \centering
    \caption{Performance of linguistic, pre-retrieval and post-retrieval predictors on VQPP. We report Pearson and Kendall correlation coefficients for Reciprocal Rank (RR) and Recall@10 (R@10) across both datasets (VATEX and MSR-VTT) and both retrieval systems (GRAM and VAST). Symbols $\dagger$ and $\ddagger$ represent statistical significance at p-values $\mathbf{0.01}$ and $\mathbf{0.001}$, respectively. Best results are in bold.}
    \label{tab:combined_results}
    \vspace{-0.25cm}
    \begin{adjustbox}{width=\textwidth}
    \setlength{\tabcolsep}{2.5pt}
    \begin{tabular}{l cc cc cc cc cc cc cc cc}
        \toprule
        \multirow{4.75}{*}{\textbf{Method}} & \multicolumn{8}{c}{\textbf{VATEX}} & \multicolumn{8}{c}{\textbf{MSR-VTT}} \\

        \cmidrule(lr){2-9}
        \cmidrule(lr){10-17}
        
         & \multicolumn{4}{c}{\textbf{GRAM}} & \multicolumn{4}{c}{\textbf{VAST}} & \multicolumn{4}{c}{\textbf{GRAM}} & \multicolumn{4}{c}{\textbf{VAST}} \\

         \cmidrule(lr){2-5}
        \cmidrule(lr){6-9}
        \cmidrule(lr){10-13}
        \cmidrule(lr){14-17}
         & \multicolumn{2}{c}{\textbf{RR}} & \multicolumn{2}{c}{\textbf{R@10}} & \multicolumn{2}{c}{\textbf{RR}} & \multicolumn{2}{c}{\textbf{R@10}} & \multicolumn{2}{c}{\textbf{RR}} & \multicolumn{2}{c}{\textbf{R@10}} & \multicolumn{2}{c}{\textbf{RR}} & \multicolumn{2}{c}{\textbf{R@10}} \\
         & Pearson & Kendall & Pearson & Kendall & Pearson & Kendall & Pearson & Kendall & Pearson & Kendall & Pearson & Kendall & Pearson & Kendall & Pearson & Kendall \\
        \midrule
        \multicolumn{17}{l}{\emph{Linguistic Baselines}} \\
        Synset Count & -0.03 & -0.01 & -0.03 & 0.00 & -0.06 & -0.04 & -0.06 & -0.05$\dagger$ & 0.17$\ddagger$ & 0.13$\ddagger$ & 0.18$\ddagger$ & 0.15$\ddagger$ & 0.14$\ddagger$ & 0.11$\ddagger$ & 0.15$\ddagger$ & 0.12$\ddagger$ \\
        Word Count & -0.05 & -0.04 & -0.08 & -0.06 & -0.03 & -0.05$\dagger$ & -0.08$\dagger$ & -0.07$\dagger$ & 0.26$\ddagger$ & 0.21$\ddagger$ & 0.27$\ddagger$ & 0.25$\ddagger$ & 0.22$\ddagger$ & 0.18$\ddagger$ & 0.22$\ddagger$ & 0.19$\ddagger$ \\
        Avg.~Word Length & -0.02 & -0.01 & 0.04 & 0.02 & 0.03 & 0.02 & 0.04 & 0.03 & -0.07 & -0.05 & -0.02 & -0.04 & -0.01 & -0.01 & 0.03 & 0.01 \\
        Avg.~Numerals & -0.04 & -0.05 & -0.06 & -0.06 & 0.04 & -0.00 & 0.03 & 0.00 & -0.02 & -0.00 & -0.05 & -0.04 & -0.04 & -0.02 & -0.02 & -0.01 \\
        Avg.~Conjunctions & -0.11$\dagger$ & -0.10$\ddagger$ & -0.07 & -0.06 & -0.00 & -0.05$\dagger$ & -0.10$\ddagger$ & -0.09$\ddagger$ & 0.08 & 0.09$\ddagger$ & 0.09$\dagger$ & 0.11$\ddagger$ & 0.04 & 0.06 & 0.04 & 0.07 \\
        Avg.~Prepositions & 0.04 & 0.02 & -0.04 & -0.03 & -0.04 & -0.04 & -0.05 & -0.04 & 0.14$\ddagger$ & 0.11$\ddagger$ & 0.16$\ddagger$ & 0.13$\ddagger$ & 0.11$\ddagger$ & 0.09$\ddagger$ & 0.09$\dagger$ & 0.07$\dagger$ \\
        \midrule
        \multicolumn{17}{l}{\emph{Pre-Retrieval Predictors}} \\
        Fine-tuned BERT & \textbf{0.29}$\ddagger$ & \textbf{0.22}$\ddagger$ & \textbf{0.30}$\ddagger$ & \textbf{0.23}$\ddagger$ & \textbf{0.30}$\ddagger$ & \textbf{0.24}$\ddagger$ & \textbf{0.31}$\ddagger$ & \textbf{0.25}$\ddagger$ & \textbf{0.40}$\ddagger$ & \textbf{0.33}$\ddagger$ & \textbf{0.41}$\ddagger$ & \textbf{0.34}$\ddagger$ & \textbf{0.39}$\ddagger$ & \textbf{0.34}$\ddagger$ & \textbf{0.41}$\ddagger$ & \textbf{0.34}$\ddagger$ \\
        Few-shot Llama-3.1-8B & 0.14$\ddagger$ & 0.09$\ddagger$ & 0.10$\dagger$ & 0.10$\dagger$ & 0.10$\dagger$ & 0.10$\ddagger$ & 0.16$\ddagger$ & 0.15$\ddagger$ & 0.28$\ddagger$ & 0.24$\ddagger$ & 0.26$\ddagger$ & 0.22$\ddagger$ & 0.25$\ddagger$ & 0.22$\ddagger$ & 0.19$\ddagger$ & 0.17$\ddagger$ \\
        \midrule
        \multicolumn{17}{l}{\emph{Post-Retrieval Predictors}} \\
        Fine-tuned CLIP & 0.10$\dagger$ & 0.11$\ddagger$ & 0.20$\ddagger$ & 0.18$\ddagger$ & 0.11$\ddagger$ & 0.13$\ddagger$ & 0.11$\ddagger$ & 0.10$\ddagger$ & 0.13$\ddagger$ & 0.23$\ddagger$ & 0.22$\ddagger$ & 0.22$\ddagger$ & 0.30$\ddagger$ & 0.31$\ddagger$ & 0.28$\ddagger$ & 0.27$\ddagger$ \\
        Correlation CNN & 0.20$\ddagger$ & 0.14$\ddagger$ & 0.21$\ddagger$ & 0.14$\ddagger$ & 0.17$\ddagger$ & 0.13$\ddagger$ & 0.27$\ddagger$ & 0.22$\ddagger$ & 0.27$\ddagger$ & 0.21$\ddagger$ & 0.33$\ddagger$ & 0.26$\ddagger$ & 0.24$\ddagger$ & 0.17$\ddagger$ & 0.27$\ddagger$ & 0.21$\ddagger$ \\
        \bottomrule
    \end{tabular}
    \end{adjustbox}
    \vspace{-0.2cm}
\end{table*}

\subsection{Post-Retrieval Predictors}
Post-retrieval methods harness the ranked list of videos returned by the system to assess how successful the search was.

\noindent\textbf{Fine-tuned CLIP.}
Instead of a simple similarity check, the fine-tuned CLIP \cite{radford2021learning} predictor is designed as a binary classifier. For each query, we consider the top 25 retrieved candidates. For each candidate video, we sample 12 frames at random and compute their embeddings with the CLIP image encoder.
We average the resulting embeddings to produce a representative video-level feature vector. This vector is concatenated with the query embedding returned by the CLIP text encoder. We then train a classification head on concatenated feature vectors representing query-video pairs, where the label is set to 1 if the candidate video is the ground-truth video, and 0 otherwise. The binary classification head consists of a 3-layer MLP with ReLU activations and high dropout to prevent overfitting. To handle class imbalance (there is only one correct pair in 25 pairs), we employ a class-weighted binary cross-entropy loss. 

At inference, query difficulty is derived from the probabilities predicted for the top 25 results. For RR, the difficulty is estimated as $1/n$, where $n$ is the rank of the first video with a probability higher than $0.5$. For Recall@10, we measure the ratio of matched (probability $> 0.5$) videos în the top 10 and top 25 retrieved results. When division by zero occurs, we set RR and Recall@10 to zero.

\noindent\textbf{Fine-tuned CLIP4Clip.}
This baseline follows the same supervised classification framework as the fine-tuned CLIP predictor, but utilizes the CLIP4Clip \cite{luo2021clip4clip} architecture as its visual backbone. Instead of simple frame averaging, we leverage the temporal transformer in CLIP4Clip to aggregate the sampled frames into a single video-level embedding, capturing motion-aware temporal dependencies. This embedding is concatenated with the query's text representation and passed through a binary classification head trained to distinguish ground-truth matches from distractors within the top 25 results. 
  
\noindent\textbf{Correlation CNN.} Sun et al.~\cite{Sun-TIP-2018} proposed to train a convolutional neural network (CNN) on correlation matrices of retrieved images. We adapt the correlation CNN to the video domain. The predictor estimates query difficulty by analyzing the visual coherence and semantic redundancy among the top-retrieved videos. For a given query, we consider the top 25 video candidates and embed each one via the CLIP image encoder. We compute the cosine similarity between each pair of video embeddings, obtaining a correlation matrix of $25\times 25$ components. The resulting correlation matrix is treated as a single-channel image and given as input to a CNN. The network comprises four conv-pooling blocks, with 64, 128, 256 and 512 filters, respectively. All layers use filters with a kernel size of $3\times 3$, activated by ReLU.  Each conv layer is followed by a max-pooling layer with a $2\times 2$ kernel and a stride of 2. The convolutional layers are followed by a global average pooling layer. The resulting features are passed through two dense layers with 512 neurons each. The CNN is trained to regress to the ground-truth query performance scores.

\subsection{Hyperparameter Configuration}
The BERT predictor is fine-tuned using the AdamW optimizer \cite{Loshchilov-ICLR-2019} for 12 epochs, with a batch size of 32 samples. The model is trained to minimize the Mean Squared Error (MSE) loss. The CLIP and CLIP4Clip predictors are fine-tuned for 25 epochs using the Adam optimizer \cite{Kingma-ICLR-1015} with a batch size of 256 samples. These predictors are trained via class-weighted binary cross-entropy. The correlation CNN is trained for 25 epochs using Adam with a batch size of 32, minimizing the MSE loss. All models are trained with early stopping. For all trained predictors, we perform grid search to select optimal values for other hyperparameters, exploring learning rates in the set $\{10^{-5}, 5\cdot 10^{-5}, 10^{-4}\}$ and weight decay values in the set $\{0, 0.01, 0.1\}$. For the LLM-based predictor (\texttt{Llama-3.1-8b-instant}), we set the temperature to 0 to obtain deterministic outputs. All experiments are carried out on a computer with
one Nvidia GeForce GTX 3090 GPU (24GB VRAM).

\section{Experiments and Results}

\subsection{Main Results} 

In Table \ref{tab:combined_results}, we report the Pearson and Kendall correlation coefficients between the predicted difficulty scores and the ground-truth performance metrics for linguistic, pre-retrieval, and post-retrieval predictors on VQPP. Results are reported across both retrieval systems and datasets. Our comparative evaluation highlights three primary trends in video QPP performance. First of all, deep pre-retrieval methods demonstrate superior performance across the board, with the fine-tuned BERT predictor consistently outperforming more complex post-retrieval models, despite not having access to the lists of retrieved videos. Second of all, we observe a significant performance drop on the VATEX dataset compared with MSR-VTT. We hypothesize that this difference exists because MSR-VTT contains significantly more descriptive and concrete queries compared with the more concise queries in VATEX. Finally, pre-retrieval performance remains largely invariant across the GRAM and VAST retrieval systems. We consider that this is caused by the fact that pre-retrieval predictors do not depend on the results retrieved by these retrieval systems, so predictors have to focus on the query content, which is the same for both models.

While post-retrieval predictors achieved superior performance in related benchmarks, such as iQPP \cite{Poesina-SIGIR-2023} and PQPP \cite{Poesina-CVPR-2025}, this does not seem to be the case for VQPP. We attribute this distinctive pattern to the fact that VATEX and MSR-VTT contain only one correct video per query, making it very hard to leverage such a weak signal from the list of candidate videos. Overall, the best performing predictor is the simple fine-tuned BERT model. Nevertheless, even the fine-tuned BERT predictor reaches correlation values that are below $0.5$, further indicating that VQPP is a challenging benchmark.

\begin{table}[t]
    \centering
    \caption{Ablation study for the number of shots ($\mathbf{k}$) used for the few-shot Llama-3.1-8B predictor. The study is conducted on the VATEX dataset and the GRAM retrieval model. Pearson and Kendall correlation coefficients are reported for the Reciprocal Rank (RR). Best results are in bold.}
    \label{tab:llama_shots}
    \vspace{-0.25cm}
    \begin{tabular}{lccc}
        \toprule
        {\textbf{Model}} & $\mathbf{k}$ & \textbf{Pearson} & \textbf{Kendall} \\
        \midrule
        Few-shot Llama-3.1-8B & $k=4$ & 0.105 & 0.061 \\
        Few-shot Llama-3.1-8B & $k=8$ & 0.123 & 0.083 \\
        Few-shot Llama-3.1-8B & $k=16$ & \textbf{0.138} & \textbf{0.094} \\
        \bottomrule
    \end{tabular}
    \vspace{-0.15cm}
\end{table}

\subsection{Ablation Studies}

\noindent\textbf{Impact of few-shot examples.}
To determine the optimal configuration for the Llama-3.1-8B predictor, we conduct a hyperparameter study on the number of in-context examples inserted in the prompt. As shown in Table~\ref{tab:llama_shots}, performance scales positively with the number of shots. Increasing $k$ from $4$ to $16$ results in higher Pearson and Kendall correlation coefficients. For the main results, we use $k=16$. Due to our limited computational resources, we did not test configurations with $k>16$. 

\begin{table}[t]
    \centering
    \caption{Comparison between CLIP and CLIP4Clip post-retrieval predictors. The study is conducted on the VATEX dataset and the GRAM retrieval model. Best results are in bold.}
    \label{tab:clip_comparison}
    \vspace{-0.25cm}
    \setlength{\tabcolsep}{3.2pt}
    \begin{tabular}{lcccc}
        \toprule
        \multirow{2}{*}{\textbf{Post-Retrieval Model}} & \multicolumn{2}{c}{\textbf{RR}} & \multicolumn{2}{c}{\textbf{Recall@10}} \\
         & Pearson & Kendall & Pearson & Kendall \\
        \midrule
        Fine-tuned CLIP4Clip & 0.040 & 0.077 & 0.047 & 0.047 \\
        Fine-tuned CLIP & \textbf{0.103} & \textbf{0.107} & \textbf{0.199} & \textbf{0.175} \\
        \bottomrule
    \end{tabular}
    \vspace{-0.2cm}
\end{table}

\noindent 
\textbf{CLIP vs.~CLIP4Clip.}
We consider two variants of CLIP-based predictor, one that uses the original model and one that uses CLIP4Clip \cite{luo2021clip4clip}, the former being designed for images and the latter for videos. Although CLIP4Clip is specifically designed for the video domain, the results reported in Table~\ref{tab:clip_comparison} indicate that the standard CLIP backbone yields superior QPP performance. For the main results reported in Table \ref{tab:combined_results}, we therefore use the original CLIP model.

\section{Application: Query Reformulation}

To demonstrate the practical utility of the VQPP benchmark beyond passive difficulty estimation, we introduce a pipeline for query reformulation. We hypothesize that a relatively-accurate QPP estimator can act as a reward model, providing a proxy objective function to guide an LLM towards generating more effective queries.

\begin{figure}[t]
  \centering
  \includegraphics[width=1.0\linewidth]{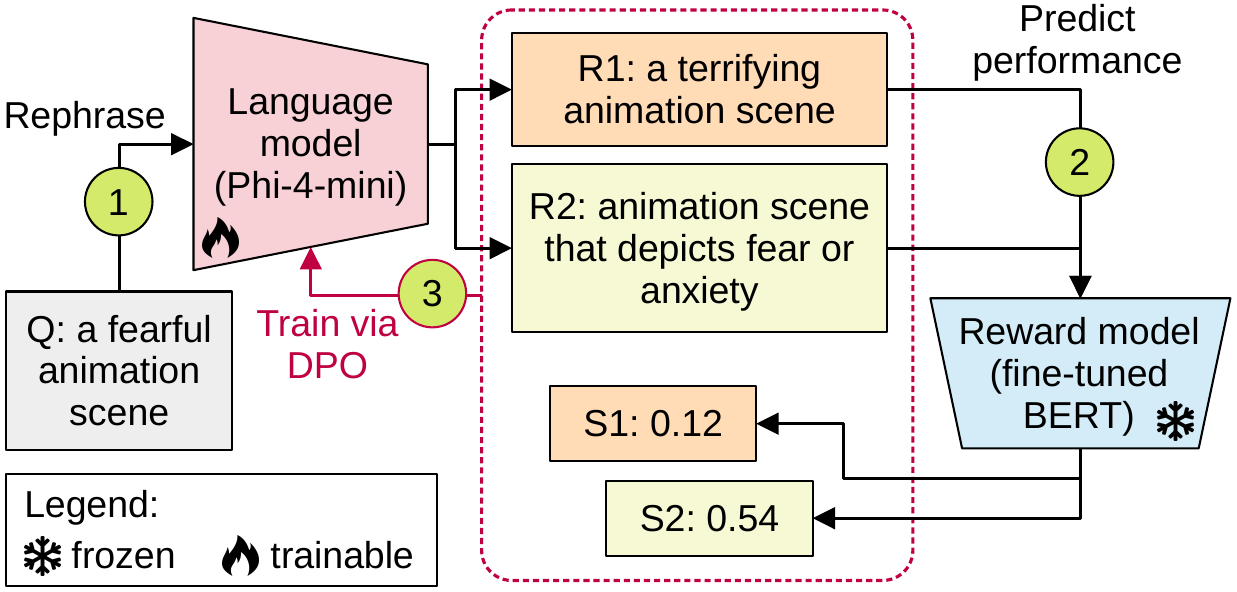}
  \vspace{-0.6cm}
  \caption{The query reformulation pipeline comprises three steps. 
  \raisebox{0.5pt}{\textcircled{\raisebox{-0.6pt}{1}}} A language model (Phi-4-mini-instruct \cite{Abouelenin-Arxiv-2025}) is prompted to provide multiple query reformulations. 
  \raisebox{0.5pt}{\textcircled{\raisebox{-0.6pt}{2}}} The resulting queries are given as input to the fine-tuned BERT pre-retrieval predictor, which scores each query in terms of retrieval performance. \raisebox{0.5pt}{\textcircled{\raisebox{-0.6pt}{3}}} Reformulated queries are arranged into (winning, losing) pairs according to the rewards given by the fine-tuned BERT predictor. The language model is optimized on the resulting pairs via Online Direct Preference Optimization (DPO) \cite{Rafailov-NeurIPS-2023,Qi-Arxiv-2024}. Best viewed in color.}
  \label{fig:dpo}
  \Description{}
  \vspace{-0.15cm}
\end{figure}

\begin{table}[t]
    \centering
    \caption{Retrieval performance (Recall@10) of the GRAM model on MSR-VTT with original vs.~rephrased queries. The queries are reformulated by Phi-4-mini, after the model is trained via DPO with rewards from fine-tuned BERT.}
    \label{tab:qref}
    \vspace{-0.25cm}
    \begin{tabular}{l c c}
        \toprule
        \textbf{Metric} & \textbf{Original} & \textbf{Rephrased} \\
        \midrule
        {Recall@10} & 47.28\% & \textbf{47.62\%} \\
        \bottomrule
    \end{tabular}
    \vspace{-0.2cm}
\end{table}

\begin{figure}[t]
  \centering
  \includegraphics[width=0.98\linewidth]{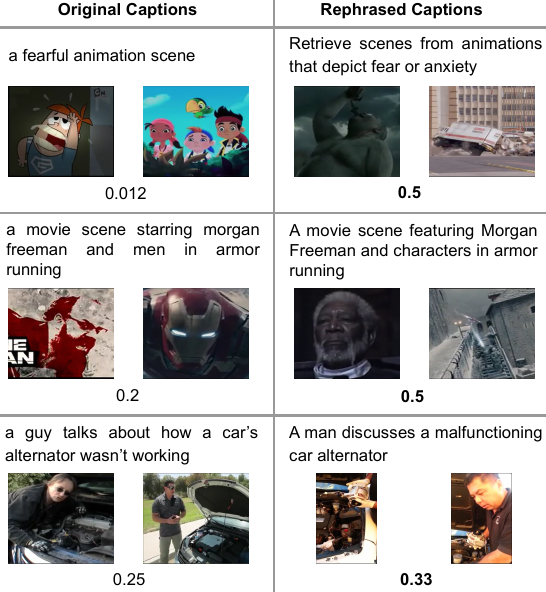}
\vspace{-0.3cm}
  \caption{Examples of original and reformulated queries, along with the top two videos retrieved from the MSR-VTT \cite{xu2016msrvtt} dataset by the GRAM model \cite{cicchetti2025gramian}. Reformulated queries obtain higher retrieval ranks (depicted in bold). Best viewed in color.}
  \label{fig:qref}
  \Description{}
  \vspace{-0.2cm}
\end{figure}

\noindent\textbf{Methodology.}
We employ Online Direct Preference Optimization (DPO) \cite{Rafailov-NeurIPS-2023,Qi-Arxiv-2024} to fine-tune an LLM for the reformulation task. The pipeline consists of two models, as illustrated in Figure \ref{fig:dpo}. The \emph{policy model}, in our case Phi-4-mini-instruct \cite{Abouelenin-Arxiv-2025}, is the base generator that learns to reformulate queries. To ensure efficient training, we fine-tune Phi-4-mini using Low-Rank Adaptation (LoRA) \cite{Hu-ICLR-2021} on all linear layers. The \emph{reward model}, in our case the fine-tuned BERT predictor, provides reward signals by judging every rephrased query. During training, the policy generates two candidate reformulations for each original query. The BERT predictor predicts difficulty scores for both. The resulting reward signals are used to determine which reformulation wins in each pair of candidate reformulations.

    
    

\noindent\textbf{Results.}
In Table~\ref{tab:qref}, we compare retrieval performance for original queries and queries reformulated by Phi-4-mini. The results indicate that Phi-4-mini can generally produce effective reformulations, leading to higher recall. The examples illustrated in Figure \ref{fig:qref} suggest that Phi-4-mini learns to restructure queries to be more descriptive and visually concrete, which explains the improved retrieval performance.

\section{Conclusion}

In this paper, we introduced VQPP, the first benchmark for QPP in the context of content-based video retrieval. VQPP comprises query performance evaluations across two video collections (VATEX and MSR-VTT) and two retrieval systems (GRAM and VAST), providing a comprehensive set of evaluation scenarios for future QPP development in CBVR. We comparatively assessed a broad range of predictors, ranging from basic linguistic features to deep pre-retrieval and post-retrieval models. Our findings indicated that VQPP is a challenging benchmark, suitable for future exploration of predictors specifically designed for CBVR.

In future work, we aim to maintain a public ranking of QPP models for CBVR and organize a challenge in an associated workshop, e.g.~\cite{Meng-ECIR-2025}, to boost the development of QPP research in CBVR.
 
\begin{acks}
This research is supported by the project ``Romanian Hub for Artificial Intelligence - HRIA'', Smart Growth, Digitization and Financial Instruments Program, 2021-2027, MySMIS no.~351416.
\end{acks}

\bibliographystyle{ACM-Reference-Format}
\bibliography{sample-base}

\end{document}